\pdfoutput=1

\documentclass[11pt]{article}

\usepackage{acl}

\usepackage{times}
\usepackage{latexsym}

\usepackage[T1]{fontenc}

\usepackage[utf8]{inputenc}

\usepackage{microtype}

\usepackage{amsmath}
\usepackage{float}
\usepackage{graphicx}
\usepackage{enumitem}
\usepackage{xcolor,colortbl}
\definecolor{Gray}{gray}{0.85}
\usepackage{multirow}
\usepackage{cleveref}
\usepackage{amssymb}
\usepackage{pifont}
\newcommand{\xmark}{\ding{55}}%
\usepackage{booktabs}
\usepackage{arydshln}

\usepackage[linecolor=orange,size=scriptsize,disable]{todonotes}

\usepackage{marginnote}

\newcommand{\seokhwan}[1]{\todo[color=blue!20] {SK: #1}}
\newcommand{\sgella}[1]{\todo[color=yellow!80] {SG: #1}}

\title{"What do others think?": \\Task-Oriented Conversational Modeling with Subjective Knowledge}

\author{Chao Zhao $^{1}$ \;  Spandana Gella $^{2}$ \; Seokhwan Kim $^{2}$ \; Di Jin $^{2}$ \; \\ \bf Devamanyu Hazarika $^{2}$ 
Alexandros Papangelis $^{2}$ \; Behnam Hedayatnia $^{2}$ \; \\ \bf Mahdi Namazifar $^{2}$ \; Yang Liu $^{2}$ \; Dilek Hakkani-Tur $^{2}$ \\
\texttt{zhaochao@cs.unc.edu}\quad \texttt{\{sgella,seokhwk,djinamzn\}@amazon.com}\\\texttt{\{dvhaz,papangea,behnam,mahdinam,yangliud,hakkanit\}@amazon.com}\\$^{1}$ UNC Chapel Hill \qquad $^{2}$ Amazon, Alexa}

\begin{document}
\maketitle
\begin{abstract}

Task-oriented Dialogue (TOD) Systems aim to build dialogue systems that assist users in accomplishing specific goals, such as booking a hotel or a restaurant. Traditional TODs rely on domain-specific APIs/DBs or external factual knowledge to generate responses, which cannot accommodate subjective user requests (e.g.,``\textit{Is the WIFI reliable?}'' or ``\textit{Does the restaurant have a good atmosphere?}''). To address this issue, we propose a novel task of subjective-knowledge-based TOD (SK-TOD). We also propose the first corresponding dataset, which contains subjective knowledge-seeking dialogue contexts and manually annotated responses grounded in subjective knowledge sources. When evaluated with existing TOD approaches, we find that this task poses new challenges such as aggregating diverse opinions from multiple knowledge snippets. We hope this task and dataset can promote further research on TOD and subjective content understanding. The code and the dataset are available at \url{https://github.com/alexa/dstc11-track5}.

\end{abstract}

\section{Introduction}

Task-oriented Dialogue (TOD) Systems aim to build dialogue systems that assist users in accomplishing specific goals, such as booking a hotel or a restaurant. 
Most  solutions of TOD are based on domain-APIs \cite{budzianowski-etal-2018-multiwoz,rastogi2020towards} and structured databases \cite{eric-etal-2017-key,wu2018globaltolocal}, which can only handle a limited range of scenarios within the scope of APIs/DBs. 
To further enlarge the model's ability of task-oriented assistance, recent works \cite{dimitrakis2018finding,kim-etal-2020-beyond,kim2021robust,feng-etal-2020-doc2dial,feng-etal-2021-multidoc2dial,majumder-etal-2022-achieving} 
incorporate unstructured textual information retrieved from the Internet into dialogue modeling. 
Most of these works focus on factual knowledge sources such as frequently asked questions (FAQs) of online products or government service guides. We refer to these models as Fact-TOD models.

\begin{figure}[t]
    \centering
    \includegraphics[width=\linewidth]{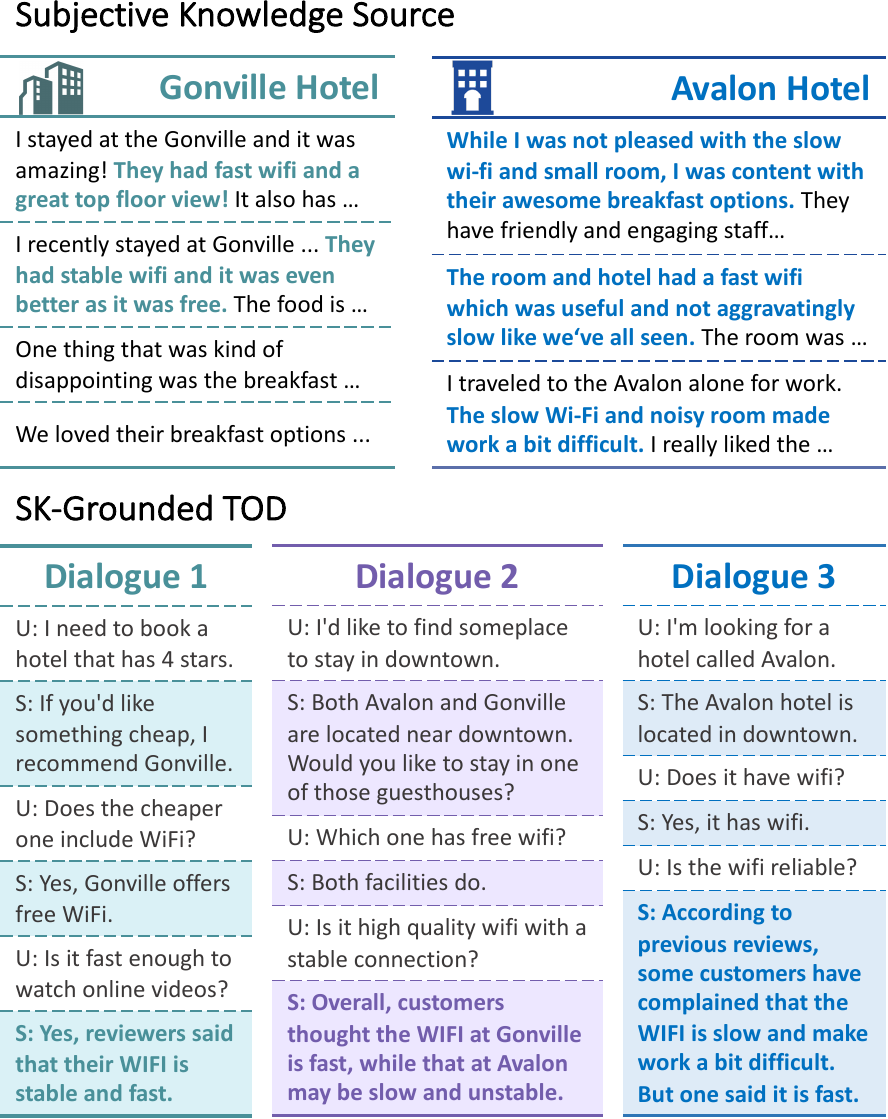}
    \caption{Examples of the SK-TOD task. The top part shows two hotels and their customer reviews. The bottom part shows three dialogue sessions between the system (denoted by S) and three users (denoted by U). The last user utterance is a subjective question about the WIFI quality of the hotel(s). The system needs to retrieve information from the relevant subjective knowledge, which is highlighted in the review text.}
    \label{fig:example}
\end{figure}

However, in many TOD tasks, users care about not only factual information but subjective insights as well, such as the experiences, opinions, and preferences of other customers. For instance, when booking a hotel or a restaurant, users often inquire about subject aspects like \textit{``Is
the WIFI reliable?''} or \textit{``Does the restaurant have a good atmosphere?''}.
To respond to such user requests, an agent needs to seek information from subjective knowledge sources, such as online customer reviews.
While subjective knowledge has been specifically studied in other NLP problems such as opinion mining \cite{liu2012survey} and question answering \cite{bjerva-etal-2020-subjqa}, incorporating it into TOD has not received significant attention.

In this work, we argue that it is important to enable the TOD model to leverage subjective knowledge for more effective task-oriented assistance. To this end, we propose a novel task of subjective-knowledge-based task-oriented dialogue (SK-TOD). SK-TOD focuses on responding to user requests that seek subjective information by incorporating user reviews as subjective knowledge. Figure \ref{fig:example} shows three examples of such requests, where customers ask about the WiFi quality of various hotels. 
User reviews are valuable resources for subjective information because even for the same aspect of a product or service, customers may have different opinions and leave either positive or negative reviews. As a result, a TOD system should consider multiple reviews to provide a comprehensive representation of user opinions. Ideally, the system's response should include both positive and negative opinions, along with their respective proportions (as exemplified in Dialogue 3). This two-sided response has been recognized as more credible and valuable for customers \cite{kamins1989two,lee2008effect,baek2012helpfulness}, thereby fostering trust in the TOD system. 

Incorporating subjective knowledge into TOD introduces two unique challenges. 
Firstly, unlike in Fact-TOD where selecting a few relevant knowledge snippets suffices, the SK-TOD model must consider all relevant knowledge snippets. 
In other words, both precision and recall matter during this process.
Secondly, the model needs to aggregate these knowledge snippets into a concise response that can faithfully reflect the diversity and proportion of opinions expressed.
Conquering these challenges requires a large-scale dataset with subjective-knowledge-grounded responses, which, to our best knowledge, is not publicly available.

To facilitate the research in subjective-knowledge-grounded TOD, we have collected a large-scale dataset, which contains 19,696 subjective knowledge-seeking dialogue contexts and manually annotated responses that are grounded on 143 entities and 1,430 reviews (8,013 sentences). 
We evaluate the performance of strong baselines on the SK-TOD task. 
Results show that there is a significant gap between human-generated and machine-generated responses, particularly in terms of the faithfulness of the sentiment proportion.
To address this issue, we propose a model that incorporates review understanding into SK-TOD. We experimentally demonstrate that responses generated by this model more effectively capture the sentiment proportion. 
Our contributions are three-fold:

\begin{itemize}
    \item We introduce a novel task of subjective-knowledge-based TOD (SK-TOD);
    \item We create and release a large-scale, human-annotated dataset designed for this task;
    \item We propose a new model and conduct extensive experiments on the proposed task.
\end{itemize}

\section{Related Work}

\subsection{Knowledge-Grounded Dialogue}
Knowledge-grounded response generation is popular in the open-domain dialogue. Numerous external knowledge sources have been explored, from structured knowledge such as fact tables \cite{DBLP:conf/emnlp/MogheABK18,liu-etal-2018-knowledge} and knowledge graphs \cite{zhang-etal-2020-grounded,moon-etal-2019-opendialkg,tuan-etal-2019-dykgchat}, to unstructured knowledge such as Wikipedia articles \cite{vougiouklis-etal-2016-neural,zhou-etal-2018-dataset,dinan2018wizard}, news articles \cite{majumder-etal-2020-interview}, web pages \cite{long2017knowledge,galley2019grounded,komeili-etal-2022-internet}, narratives \cite{xu2021enhancing,gopalakrishnan2019topical}, user reviews and comments \cite{DBLP:conf/emnlp/MogheABK18,Ghazvininejad2018AKN}, and so on. 
Grounding on external knowledge makes the response more informative and meaningful when compared with models that solely rely on the dialog context. 

Regarding task-oriented dialogues, previous works have primarily focused on domain-specific APIs and databases to support the dialogue response \cite{levin2000stochastic,singh2002optimizing,williams2007partially, eric-etal-2017-key,wu2018globaltolocal}, which can only support a limited scope of user queries. Later works ground task-oriented dialogues to web pages \cite{penha2019introducing,chen-etal-2022-ketod}, government service documents \cite{saeidi-etal-2018-interpretation,feng-etal-2020-doc2dial,feng-etal-2021-multidoc2dial}, and FAQ knowledge snippets \cite{kim-etal-2020-beyond,kim2021robust}. Different from these works where factual knowledge is utilized, we apply subjective knowledge to generate the response and ground in multiple knowledge snippets. While \citet{majumder-etal-2022-achieving} also explored grounding TOD in user reviews, they did not consider the diversity of opinions.

\begin{table*}[t]
    \centering
    
\small
\centering
\begin{tabular}{lccccccccc}
\hline
    ~ & Size & Manual & Dial & TOD & Query & Aspect & Senti & Mul-Knwl & Senti-\%  \\ \hline
    Semeval/MAMS \shortcite{pontiki2016semeval,jiang-etal-2019-challenge} & 5K/22K & $\checkmark$ & \xmark & n/a & \xmark & $\checkmark$ & $\checkmark$ & \xmark & n/a  \\ 
    Space \shortcite{Angelidis2021ExtractiveOS} & 1K & $\checkmark$ & \xmark & n/a & \xmark & $\checkmark$ & $\checkmark$ & $\checkmark$ & \xmark  \\ 
    Yelp/Amazon \shortcite{chu2019meansum,Brazinskas2020UnsupervisedOS} & 200/180 & $\checkmark$ & \xmark & n/a & \xmark & \xmark & $\checkmark$ & $\checkmark$ & \xmark  \\ 
    Justify-Rec \shortcite{ni2019justifying} & 1.3M & \xmark & \xmark & n/a & \xmark & $\checkmark$ & \xmark & $\checkmark$ & \xmark  \\ 
    AmazonQA \shortcite{mcauley2016addressing} & 309K & \xmark & \xmark & n/a & $\checkmark$ & \xmark & \xmark & \xmark & n/a  \\ 
    SubjQA \shortcite{bjerva-etal-2020-subjqa} & 10K & \xmark & \xmark & n/a & $\checkmark$ & $\checkmark$ & $\checkmark$ & \xmark & n/a  \\ 
    \hline
    Holl-E \shortcite{DBLP:conf/emnlp/MogheABK18} & 9K & $\checkmark$ & $\checkmark$ & \xmark & \xmark & \xmark & \xmark & $\checkmark$ & \xmark  \\ 
    Foursquare \shortcite{Ghazvininejad2018AKN} & 1M & \xmark & $\checkmark$ & \xmark & \xmark & \xmark & \xmark & $\checkmark$ & n/a  \\ \hline
    SK-TOD (Ours) & 20K & $\checkmark$ & $\checkmark$ & $\checkmark$ & $\checkmark$ & $\checkmark$ & $\checkmark$ & $\checkmark$ & $\checkmark$ \\ \hline
\end{tabular}

    \caption{\label{tab:related}Comparison between SK-TOD and other benchmarks based on the subjective content. We consider if the dataset is manually annotated, dialogue-based, task-oriented, and query-focused. We also list if it considers  aspect and sentiment, multiple knowledge snippets (Mul-Knwl), and the proportion of two-sided sentiments (Senti-\%). }
\end{table*}

\subsection{Subjective Content Understanding}
Besides being used as external knowledge sources in dialogue systems, subjective content, especially user reviews, has been studied in various non-conversational NLP tasks. For example, opinion mining \cite{pontiki2016semeval,jiang-etal-2019-challenge} focuses on extracting opinions and sentiments from user reviews. Opinion summarization \cite{chu2019meansum,zhao2020weakly,Brazinskas2020UnsupervisedOS,Angelidis2021ExtractiveOS} is used to distill multiple opinions into concise summaries. Subjective question answering \cite{mcauley2016addressing,bjerva-etal-2020-subjqa} have been proposed to answer questions based on user reviews.
Explainable recommendation \cite{ni2019justifying} aims to generate review-based explanations for the items recommended by a recommendation system. 
Table \ref{tab:related} provides detailed comparisons between SK-TOD and these subjective-content-based benchmarks. Generally, SK-TOD requires creating a response that is appropriate to the dialogue context. It also requires grounding in multiple subjective knowledge and explicitly considers the diversity of opinions and the proportion of sentiments. 

\seokhwan{let's describe other subproblems like KTD, entity tracking and KS in addition to RG.}

\sgella{we don't mention about lack of work or that our work does emphasize on having aspects for user queries and reflecting sentiment distribution of knowledge snippets in response which was not studied earlier.  Other works did parts of this (for example opinion memorization: refer those). Some pointers added below.}

\section{Problem Formulation}

Formally, we have a dialogue context $C=[U_1, S_1, U_2, S_2, \cdots, U_t]$ between a user and a system, where each user utterance $U_i$ is followed by a system response utterance $S_i$, except for the last user utterance $U_t$. The dialogue involves one or more entities, denoted as $\mathcal{E}=\{e_1, \cdots, e_m\}$. Alongside the dialogue, we have a subjective knowledge source $\mathcal{B}=\{(e_1, \mathcal{R}_1), (e_2, \mathcal{R}_2), \cdots\}$ containing all the entities and their corresponding customer reviews. Each entity $e$ is associated with multiple reviews $\mathcal{R}=\{R_1, R_2, \cdots\}$. Each review can be divided into segments $[K_1, K_2, \cdots ]$, such as paragraphs, sentences, or sub-sentential units. In this work, we regard each review sentence as a knowledge snippet. 

The SK-TOD task aims to identify whether $U_t$ is a subjective knowledge-seeking request and, if it is, to select the relevant knowledge snippets $\mathcal{K}^+$ from the knowledge source and finally generate a response $S_t$ grounded on $\mathcal{K}^+$.

\section{Data Collection and Statistics}

We ground the data collection in MultiWOZ \cite{budzianowski-etal-2018-multiwoz,eric-etal-2020-multiwoz}. We select dialogues from the domains of hotels and restaurants. 
The data collection is conducted by a group of crowd workers through Amazon Mechanical Turk (AMT). 
To control the data quality, we only choose workers that are pre-qualified. More details can be found in Appendix \ref{app:data}.

\subsection{Annotation Guideline}

Dialogues in MultiWOZ are collected based on single or multiple entities as the back-end database. %
To create a subjective knowledge source to support the SK-TOD task, we first collect multiple user reviews for each entity. 
To control the review collection, we provide the reviewer's persona, as well as the aspects and sentiments of reviews to workers. 
We then ask workers to write a review with all the given information included. 
After collecting the reviews, we also annotate the aspect and sentiment information for each review sentence. Overall, we select 33 hotels and 110 restaurants from MultiWOZ, and collect 10 reviews for each entity. On average, each review contains 5.6 sentences and 56.71 tokens. More details about the review collection can be found in \Cref{app:data}.

After obtaining the reviews, we go back to the dialogue data to create the subjective user request. Following a similar procedure in \citet{kim-etal-2020-beyond}, for each dialogue, we provide an aspect that users are interested in (e.g., WIFI-quality of the hotel) and then ask the worker to insert a subjective user request into the dialogue. Workers are requested to carefully select the insertion position and write an utterance to maintain coherence and naturalness in the dialogue flow. Finally, we use the partial dialog until this newly inserted turn as an instance in our data. Utterances that come after the insertion position are removed from the dialogue instance. %

So far, we've collected the dialogue context $C$ and the subjective knowledge source $\mathcal{B}$. 
The final step is to ground the dialogue in the knowledge source.
We first ask workers to identify entities that are relevant to the subjective user request as gold entities. 
We then align the user request and review sentences of the gold entities by matching their aspect. 
For example, if the aspect of a user request is about the ``WIFI quality'' of a hotel, all review sentences discussing the ``WIFI quality'' of that specific hotel will be considered relevant knowledge snippets.
\footnote{Note that the aspect information is only used to build the dataset but is not included in the problem formulation of SK-TOD, which means it is not available for model training. The goal of SK-TOD is to handle user requests with arbitrary aspects, and therefore we do not define a taxonomy of aspects in the task like what is done in dialogue state tracking.}
Finally, we provide the dialogue context $C$ and all related knowledge snippets $\mathcal{K}^+$ and ask workers to generate a natural and faithful response. We explicitly instruct workers to consider the diversity and proportion of opinions in all relevant knowledge snippets during response creation. Detailed instructions can be found in Appendix \ref{app:data}.

\subsection{Quality Control}
To ensure the quality of our dataset, we took great care in selecting pre-qualified workers and designing annotation interfaces. We further conducted a human verification task on the entire dataset to identify invalid instances. The annotation showed that 81.89\% of subjective-knowledge-seeking user turns are valid, with an Inter-Annotator Agreement (IAA) score of 0.9369 in Gwet's gamma. For agent response turns, 96.78\% were  valid, with an IAA score of 0.9497 in Gwet's gamma. Any invalid instances were filtered out or manually corrected before finalizing the dataset. We paid workers an average of \$13.82/hr for data annotation and \$14.77/hr for data verification. Both exceed the local living minimum wage. The details of our payment settings are elaborated on in \Cref{app:data}.

\subsection{Data Statistics}

\begin{table}[t]
    \centering
    \small
    \centering
    \begin{tabular}{lrrr}
    \toprule
        ~ & \textbf{Train} & \textbf{Val} & \textbf{Test}  \\ \midrule
        \# instances & 14768 & 2129 & 2799  \\ 
        \# seen instances & 14768 & 1471 & 1547  \\ 
        \# unseen instances & 0 & 658 & 1252  \\ 
        \# multi-entity instances & 412 & 199 & 436  \\ \midrule
        Knowledge Snippets &&& \\
        Avg. \# snippets per instance & 3.80 & 4.07 & 4.21  \\
        Avg. \# tokens per snippet & 14.68 & 15.49 & 14.5  \\ \midrule
        Dialogue &&& \\
        Avg. \# uttrances per instance & 9.29 & 9.44 & 9.36  \\
        Avg. \# tokens per request & 8.65 & 8.94 & 9.12 \\
        Avg. \# tokens per response & 24.18 & 23.61 & 23.86 \\ \bottomrule
    \end{tabular}

    \caption{\label{tab:stat}Basic statistics of our dataset. }
\end{table}

We collected a total of 19,696 instances consisting of subjective user requests and subjective-knowledge-grounded responses. The average length of the subjective user request and the agent response is 8.75 and 24.07 tokens, respectively. While most of the instances contain a single entity, there are 1,047 instances where multiple entities are compared (like Dialogue 2 in Figure \ref{fig:example}). On average, each instance requires 3.88 subjective knowledge snippets. To help identify the subjective knowledge-seeking user request, we also randomly sample another 18,383 dialogues with non-subjective user requests from the original MultiWOZ dataset.

We split the dataset into training (75\%), validation (10.8\%), and test (14.2\%) sets. Table \ref{tab:stat} presents the detailed statistics of each subset. Both the validation and test sets contain two subsets: the \textit{seen} subset where the aspects of these instances are included in the training set, and the \textit{unseen} subset where the aspects are not included in the training set. The unseen subset is designed to evaluate models’ ability to generalize to arbitrary aspects.

\section{Subjective-Knowledge-Grounded TOD}

\begin{figure}[t]
    \centering
    \includegraphics[width=1\linewidth]{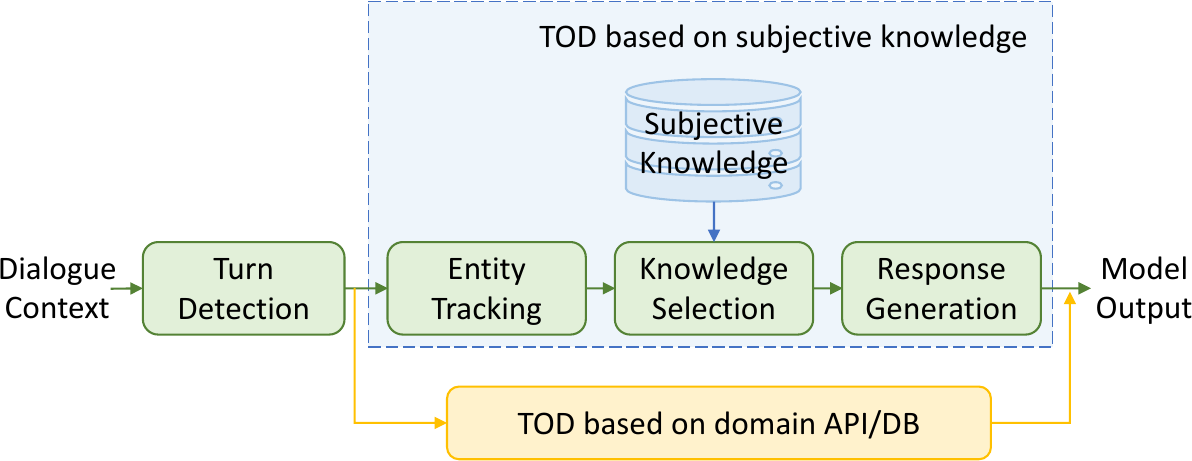}
    \caption{The pipeline architecture of SK-TOD.  } %
    \label{fig:arch}
\end{figure}

In this section, we describe the method for SK-TOD. 
As shown in Figure \ref{fig:arch}, we follow the pipeline introduced by \citet{kim-etal-2020-beyond} which comprises four sequential sub-tasks: knowledge-seeking turn detection (KTD), entity tracking (ET), knowledge selection (KS), and response generation (RG). We elaborate on each subtask below.

\subsection{Knowledge-Seeking Turn Detection}
The goal of KTD is to identify the user request that requires subjective knowledge. 
We regard it as a binary classification problem, %
where the input is the dialogue context $C$ and the output is a binary indicator.

We employ a pre-trained language model (e.g., BERT \cite{devlin-etal-2019-bert}) to encode $C$ and adopt the hidden state of the first token as its representation. 
Then we apply a classifier to obtain the probability that the current user request is seeking subjective knowledge. That is, 
\begin{equation}
\begin{aligned}
    h &= \text{Enc}(C)\\
    P(C) &= \text{softmax} \left(\text{FFN} \left( h \right) \right).
\end{aligned}
\end{equation}

The model is finetuned with the binary cross-entropy loss.

\subsection{Entity Tracking}
The goal of ET is to identify the entities $\mathcal{E}=\{e_1, \cdots, e_m\}$ that are relevant to the user request. It can help to reduce the number of candidates during the knowledge selection step.

We adopt a word-matching-based method used by \citet{jin2021can} to extract relevant entities. It first normalizes entity names in the knowledge source using a set of heuristic rules. 
Then a fuzzy n-gram matching is performed between the normalized entity and all dialogue turns. 
To find the entities that are relevant to the last user request, we choose the last dialogue turn in which the entities are detected and use these entities as the output of ET. We leave the tracking of aspects being questioned over multiple turns as future work.

\subsection{Knowledge Selection}
The goal of KS is to select the knowledge snippets that are relevant to the user's request. 
The inputs are the dialogue context $C$ and a set of knowledge snippets candidates $\mathcal{K}$, which is a combination of all knowledge snippets of the relevant entities in $\mathcal{E}$. The output $\mathcal{K}^+ \subseteq \mathcal{K}$ is a subset of relevant knowledge candidates.  Note that there might be multiple knowledge snippets in $\mathcal{K}^+$. 

To select relevant knowledge snippets, we calculate the relevance score between the dialogue context $C$ and a knowledge snippet $K\in \mathcal{K}$. 
We regard it as a pairwise text scoring problem and consider two popular approaches: bi-encoder \cite{mazare-etal-2018-training} and cross-encoder \cite{wolf2019transfertransfo}. 
Generally, the bi-encoder approach is more efficient while the cross-encoder approach is more accurate.

For the bi-encoder approach, we encode $C$ and $K$ separately using the same pre-trained encoder and obtain two representations, $h_C$ and $h_K$. Following \citet{reimers-gurevych-2019-sentence}, we use the concatenation of $h_C$, $h_K$, and $|h_C - h_K|$ as features and apply a classifier to obtain the probability of relevance. 
That is,

\begin{equation}
\begin{aligned}
    h_C &= \text{Enc}(C),\quad h_K = \text{Enc}(K)\\
    P(C, K) &= \text{softmax} \left(\text{FFN} \left(h_c, h_K, |h_C - h_K| \right) \right).
\end{aligned}
\end{equation}

For the cross-encoder approach, we encode the concatenation of $C$ and $K$ to obtain a contextualized representation. That is, 
\begin{equation}
\begin{aligned}
    h &= \text{Enc}(C, K)\\
    P(C, K) &= \text{softmax} \left(\text{FFN} \left( h \right) \right).
\end{aligned}
\end{equation}

During training, we use all relevant knowledge snippets to construct positive ($C$, $K$) pairs. 
Due to the large number of irrelevant knowledge snippets, we randomly sample the same number of irrelevant snippets to form negative pairs. We optimize the model using the binary cross-entropy loss. During inference, we predict the relevance probability for all knowledge snippets in the candidates. Since both precision and recall are crucial in KS, instead of selecting the top few results,  we use a threshold, estimated from the validation set, to determine the relevancy of each knowledge snippet.

\subsection{Response Generation}
The goal of RG is to create an utterance $S_t$ that addresses the user's request. This response is generated based on the dialogue context $C$ and the set of relevant knowledge snippets $\mathcal{K}^+$. To accomplish this, we concatenate $\mathcal{K}^+$ and $C$ as the input and use a pre-trained generation model to generate the response. 
We consider both the decoder-only model, such as GPT-2 \cite{radford2019language}, and the encoder-decoder model, such as BART \cite{lewis2020bart}.
The model is trained to maximize the generation probability $p(S_T \mid C, \mathcal{K}^+)$. %

To accurately capture the diversity and proportion of opinions, the model needs to understand the sentiment polarity of each knowledge snippet, which is challenging due to the lack of direct supervision. To address this issue, we apply a state-of-the-art aspect-based sentiment analysis (ABSA) model \cite{zhang-etal-2021-aspect-sentiment} to predict the sentiment $Z=[z_1, \cdots, z_i, \cdots]$ for each knowledge snippet $K_i \in \mathcal{K}^+$. Then we incorporate the sentiment information into RG by maximizing $p(S_T \mid C, \mathcal{K}^+, Z)$. 

More specifically, we first convert the predicted $z_i$ into a natural language description using templates, and then append it to the end of the corresponding $K_i$ as the enhanced input of RG. For example, given the knowledge snippet as ``\textit{The ambience was so fun.}'', the ABSA model detects the aspect-based sentiment as (``ambience'', ``positive''). We first convert the sentiment into a natural language ``\textit{ambience is great.}'' and then enhance the knowledge snippet as ``\textit{The ambience was so fun. ambience is great.}''. We refer to Appendix \ref{app:absa} for more details.

\section{Experiments on Sub-Tasks}
We first conduct experiments on each individual subtask.
To avoid any error accumulation from upstream tasks, we use the gold output of the previous task as the input to the current target task. 
The detailed experimental setup can be found in Appendix \ref{app:training}.

\subsection{Knowledge-Seeking Turn Detection}
\noindent \textbf{Setting}\quad
We conduct experiments using various pre-trained language models, including BERT \footnote{We use the base version of all pre-trained models.} \cite{devlin-etal-2019-bert}, RoBERTa \cite{liu2019roberta}, ALBERT \cite{Lan2020ALBERT:}, and DeBERTa \cite{he2021deberta}. 

\vspace{6pt}
\noindent \textbf{Evaluation}\quad
We report the precision, recall, $F_1$ score, and accuracy score.

\begin{table}[t]
    \centering
    \small
    \setlength{\tabcolsep}{10pt}
    \renewcommand{\arraystretch}{1.2}
\begin{tabular}{lcccc}
    \toprule
        ~ & \textbf{Acc} & \textbf{P} & \textbf{R} & \textbf{F}  \\ \midrule
        BERT & 99.67 & 99.75 & 99.61 & 99.68  \\ %
        RoBERTa & 99.74 & 99.86 & 99.64 & 99.75  \\ %
        ALBERT & 99.49 & 99.64 & 99.36 & 99.50  \\ %
        DeBERTa & 99.71 & 99.86 & 99.57 & 99.71 \\ \bottomrule
    \end{tabular}

    \caption{\label{tab:ktd_results}Results of KTD task. Models are evaluated using Accuracy, Precision, Recall, and $F_1$. All models achieve similar and near-perfect performance. }
\end{table}

\vspace{6pt}
\noindent \textbf{Results}\quad
Table \ref{tab:ktd_results} shows the results of the KTD task. All models achieve similar and near-perfect performance, which is in line with the findings of \citet{kim-etal-2020-beyond}. \sgella{these seem very high compared to FAQ task. May be we should add some explanation for this.}
It demonstrates that it is feasible to identify the user requests that require subjective knowledge, allowing them to be explicitly addressed by an SK-TOD component. However, this KTD classifier's performance may be specific to this dataset or similar domains, and its generalizability to unseen domains or knowledge types requires further exploration in future works. 

\subsection{Entity Tracking}

\noindent \textbf{Setting}\quad We follow the setting of \citet{jin2021can} to run the ET method.

\vspace{6pt}
\noindent \textbf{Evaluation}\quad
We report the instance-level accuracy score. An instance is regarded as accurate only if the predicted entities match exactly with the gold entities. %

\vspace{6pt}
\noindent \textbf{Results}\quad
The fuzzy n-gram matching method achieves an instance-level accuracy of 92.18\%. We further analyzed the type of errors. For 1.8\% of the instances, there is at least one gold entity missing from the predicted entities. For 7.6\% of the instances, the predicted entities contain at least one spurious entity. The latter error case can be further reduced by using model-based matching approaches, which we leave as future work.    

\subsection{Knowledge Selection}

\noindent \textbf{Setting}\quad
We fine-tune the KS models following the same setting as in the KTD task. Additionally, we compare them with traditional information retrieval (IR) baselines, such as TF-IDF \cite{manning2008introduction} and BM25 \cite{robertson2009probabilistic}. 

\vspace{6pt}
\noindent \textbf{Evaluation}\quad
Knowledge selection can be viewed as either a classification task or a retrieval task. 
For classification, we use precision, recall, and $F_1$ measures. We calculate these measures at both the instance level and the snippet level. 
For the instance level, we first calculate $P$/$R$/$F_1$ for each instance, and then take the average over all instances as the final scores.
For the snippet level, instead of computing $P$/$R$/$F_1$ for each instance, we calculate these scores for all <$C$, $K$> pairs in the entire dataset. 
Regarding retrieval evaluation, we use mean-average-precision (mAP) as the metric, which is not dependent on a specific threshold value and can reflect the overall ranking positions of all relevant knowledge snippets. 
Since the total number of the relevant knowledge snippets can vary for each instance, we do not include top-K-based measures like Precision@K or Recall@K, which are commonly used in other Fact-TOD and knowledge-grounded open-domain dialogue tasks.

\begin{table}[t]
    \centering
    \small
\renewcommand{\arraystretch}{1.2}
\setlength\tabcolsep{1.5pt} %
    \begin{tabular}{l cccc ccc c}
    \toprule

     &
     \multicolumn{3}{c}{\textbf{Instance-level}} &&  \multicolumn{3}{c}{\textbf{Snippet-level}} & \multirow{2}{*}{\textbf{mAP}}\\
    \cline{2-4} \cline{6-8}

        ~ & \textbf{P} & \textbf{R} & \textbf{F} && \textbf{P} & \textbf{R} & \textbf{F} &   \\ \midrule
         \textit{IR Baselines} &&&&&&&& \\
        TF-IDF & 34.61 & 70.33 & 40.46 && 23.81 & 65.00 & 34.85 & 45.97  \\ 
        BM25 & 31.38 & 40.95 & 32.21 && 31.14 & 32.42 & 31.77 & 45.42 \\ \midrule
        \textit{Bi-encoder} &&&&&&&& \\
        BERT & 56.66 & 70.06 & 59.31 && 58.87 & 74.69 & 65.84 & 71.59  \\ 
        RoBERTa & 60.98 & 83.06 & 66.47 && 54.40 & \textbf{85.38} & 66.46 & 77.25  \\ 
        ALBERT & 70.21 & 78.74 & 70.43 && \textbf{63.13} & 78.90 & 70.14 & 81.62  \\ 
        DeBERTa & \textbf{71.46} & \textbf{83.18} & \textbf{72.44} && 62.64 & 83.50 & \textbf{71.58} & \textbf{83.43} \\ \midrule
        \textit{Cross-encoder} &&&&&&&& \\
        BERT & 85.18 & 86.01 & 83.33 && 82.40 & 83.82 & 83.11 & 90.06  \\ 
        RoBERTa & 81.59 & 83.62 & 80.53 && 82.20 & 80.77 & 81.48 & 88.98  \\ 
        ALBERT & \textbf{86.18} & 87.29 & 84.22 && \textbf{83.56} & 84.78 & 84.16 & 90.50  \\ 
        DeBERTa & 86.07 & \textbf{87.64} & \textbf{84.6} && 82.70 & 85.71 & \textbf{84.18} & \textbf{91.84} \\
        \quad \textsc{Seen} & 88.80 & 93.45 & 89.93 && 90.83 & 89.90 & 90.37 & 95.70 \\
        \quad \textsc{Unseen} & 82.68 & 80.47 & 78.03 && 69.98 & 78.29 & 73.90 & 87.07 \\ \bottomrule
    \end{tabular}

    \caption{\label{tab:ks_results}Results of the KS task. Models are evaluated using instance-level and snippet-level classification measures, as well as mAP, a retrieval-based measure. DeBERTa achieves the best performance among all evaluation measures.}
\end{table}

\vspace{6pt}
\noindent \textbf{Results}\quad
Table \ref{tab:ks_results} shows the results of the KS task. Firstly, when comparing our models with IR baselines, all of the trained models outperform the baselines, indicating that the KS model can benefit from the annotated training data. 
We then compare bi-encoder models and cross-encoder models, and as expected, cross-encoder models outperform bi-encoder models by a large margin.
When comparing the performance of different pre-trained models, there is a notable difference among the models under the bi-encoder setting. The variance becomes smaller when applying the cross-encoder architecture. DeBERTa achieves the best performance on all measures in both the bi-encoder and cross-encoder settings. %

Finally, we compare the performance between the seen subset and the unseen subset. 
At the bottom of Table \ref{tab:ks_results}, we list the performance of DeBERTa on both the seen and unseen test subsets. The results reveal a large gap between the performance of the two subsets, indicating that one of the challenges for the KS model is to generalize from seen aspects to unseen aspects. 

\subsection{Response Generation}

\noindent \textbf{Setting}\quad
we experiment with decoder-only generation models such as GPT-2 \cite{radford2019language} \footnote{We use the base-version of all pre-trained models.} and DialoGPT \cite{zhang2020dialogpt}, as well as encoder-decoder models such as BART \cite{lewis2020bart} and T5 \cite{raffel2020exploring}. We also include two ABSA-enhanced models, namely BART$_\text{ABSA}$ and T5$_\text{ABSA}$. During decoding, we use beam-search with top-K sampling \cite{fan-etal-2018-hierarchical}. We set the beam size as 5 and sample from the top 50 tokens. We also compare with a random extractive baseline (EXT), where the response is created by randomly selecting a relevant knowledge snippet. 

\vspace{6pt}
\noindent \textbf{Evaluation}\quad
Following the evaluation of other generation tasks, We employ several automatic evaluation metrics, including BLEU \cite{DBLP:conf/acl/PapineniRWZ02},
ROUGE \cite{lin-2004-rouge}, METEOR \cite{DBLP:conf/acl/BanerjeeL05}, as well as BERTScore \cite{bert-score}, to evaluate the quality of the generated responses compared to the reference responses. 
We also conduct a human evaluation, where we ask crowd workers to evaluate the quality of responses.

\vspace{6pt}
\noindent \textbf{Results}\quad
\begin{table}[t]
    \centering
    \small
\centering
\renewcommand{\arraystretch}{1.2}
\setlength\tabcolsep{2pt} %
\begin{tabular}{lccccccc}
\toprule
    ~ & \textbf{BLEU} & \textbf{R-1} & \textbf{R-2} & \textbf{R-L} & \textbf{MT} & \textbf{BS} & \textbf{Len}  \\ \midrule
    EXT & 2.89 & 23.17 & 6.53 & 18.33 & 9.62 & 30.83 & 14.93 \\ \midrule
    GPT2 & 9.04 & 33.9 & 13.52 & 26.73 & 16.27 & 39.73 &  22.66  \\ 
    DialoGPT & 9.19 & 33.6 & 13.62 & 26.81 & 16.15 & 39.72 & 22.05  \\ \midrule
    BART & 10.8 & 36.35 & 15.04 & 28.57 & 17.96 & 41.12 & 24.02  \\ 
    BART$_\text{ABSA}$ & 10.78 & 36.30 & 15.36 & 28.47 & 18.06 & \textbf{41.75} & 23.66 \\
    T5 & 10.72 & 36.50 & \textbf{15.57} & 28.81 & \textbf{18.33} & 40.84 & 25.36 \\ 
    T5$_\text{ABSA}$ & \textbf{10.97} & \textbf{36.66} & 15.51 & \textbf{28.88} & 18.15 & 40.94 & 24.75 \\\bottomrule
\end{tabular}

    \caption{\label{tab:rg_results}Results of RG task. Models are evaluated using BLEU, ROUGE (R-1, R-2, R-L), METEOR (MT), and BertScore (BS). We also listed the average length (Len) of the generated response. Encoder-decoder models such as BART and T5 achieve better performance compared with GPT2-based models. }
\end{table}
As presented in Table \ref{tab:rg_results}, machine-generated responses significantly outperform the extractive responses. Encoder-decoder models achieve better performance across all automatic measures compared to GPT-based models, indicating that they are more suitable for this task. \sgella{I think this claim would look different if we train models on FAQ test on reviews. I remember initial experiments showed that GPT based models were doing better than BART/T5 in those cases.} 
They also tend to generate longer responses. There is no clear difference in automatic measures when comparing BART and T5. For ABSA-enhanced models, BART$_\text{ABSA}$ achieves the best performance on BertScore, while T5$_\text{ABSA}$ achieves the best score on BLEU and ROUGE. 

\vspace{6pt}
\noindent \textbf{Human Evaluation}\quad
To obtain a more reliable assessment of response quality, we also conduct a human evaluation on AMT. We use the same group of workers involved in the data collection process. During the evaluation, we show the dialogue context, the oracle knowledge snippets, and all responses (both the reference and the generated responses) to the workers. 
We randomly sample 240 instances from the test set for evaluation. For each instance, we ask three independent workers to compare the responses based on three measures:
\begin{itemize}\itemsep0em 
    \item Appropriateness: whether the response is fluent and naturally connected to the dialogue context.
    \item Aspect Accuracy: whether the response provides relevant and useful information to the aspect that the user queried.
    \item Sentiment Accuracy: whether the sentiment proportion provided by the response is consistent with that of the subjective knowledge.
\end{itemize}

For sentiment accuracy, we first ask workers to annotate the sentiment label of each knowledge snippet, and then evaluate each response. All three measures are evaluated using a 5-Point Likert scale. The system-level score is computed as the average score over all instances and workers for each system.
The compensation for workers was set at \$0.25 for the tasks of appropriateness and aspect accuracy, and \$0.4 for the task of sentiment accuracy. The average hourly pay for the crowd workers was \$15.25/hr, \$14.40/hr, and \$14.85/hr for each evaluation task, exceeding the local living minimum wage. %

Table \ref{tab:human_results} shows the results of human evaluation for response generation. The inter-annotator agreement scores for each task are 0.7270, 0.7535, and 0.6239 in Gwet’s gamma, respectively. The results show that machine-generated responses are comparable to the references in terms of appropriateness and aspect accuracy. Moreover, incorporating ABSA can improve the model's performance in sentiment accuracy. However, there is still a large gap in sentiment accuracy between the best model-generated responses and the references, indicating that faithfully aggregating sentiment information from multiple knowledge snippets is still a challenging task for current models.

\begin{table}[t]
    \centering
    \small
\centering
    \begin{tabular}{lccc}
    \toprule
        ~ & \textbf{Approp.} & \textbf{Asp-Acc} & \textbf{Senti-Acc}  \\ \midrule
        EXT & 2.65 & 3.32 & 3.13  \\ 
        GPT2 & 4.55 & 4.54 & 3.20  \\ 
        BART & 4.55 & 4.68 & 3.56  \\ 
        BART$_\text{ABSA}$ & 4.58 & 4.66 & 3.80  \\ 
        T5 & 4.40 & 4.63 & 3.87  \\
        T5$_\text{ABSA}$ & 4.49 & 4.67 & 3.98  \\ \midrule
        Reference & 4.70 & 4.77 & 4.50 \\ \bottomrule
    \end{tabular}
    \caption{\label{tab:human_results}Results of human evaluation for RG. }
\end{table}

\begin{table}[t]
    \centering
    \scriptsize
    \centering
    \begin{tabular}{p{0.95\linewidth}}
    \toprule
    \textbf{Dialogue History:}\\
        U: Are there any hotels (not guesthouses) that have 0 stars?\\
        S: Yes, actually. Cityroomz is a moderately priced 0 star hotel.\\
        U: Do they have free wifi?\\
        S: Yes, they do have free wifi! Would you like me to book a stay for you?\\
        U: Does the Cityroomz have strong water pressure in the shower?\\\noalign{\vskip 0.75ex}\hdashline\noalign{\vskip 0.75ex}
    \textbf{Reviews:}\\
        Review 1: However, I noticed that when taking a shower there was very little water pressure which made the shower disappointing.\\
        Review 2: The water pressure is not good and it comes out really slow.\\
        Review 3: Everything else was a disaster, especially the lack of cleanliness in the room when we checked in, and the incredibly low water pressure in the bathroom when we took our morning showers.\\
        Review 4: I encounter a problem when trying to shower and washing hands.\\
        Review 5: There is no luggage storage service or anything like that, and the water pressure is much to be desired.\\\noalign{\vskip 0.75ex}\hdashline\noalign{\vskip 0.75ex}
    \textbf{Model Outputs:}\\
        \textbf{GPT-2:} The Cityroomz gets some mixed reviews on their water pressure. Some guests found the water pressure to be very good while others found it to be disappointing. Would you like to know more about them?\\
        \textbf{BART:} The Cityroomz hotel has very low water pressure according to most guests. Only one person had a good experience with it.\\
        \textbf{BART$_\text{ABSA}$:} The Cityroomz has a lot of complaints about their water pressure being a bit low. Would you like me to find another place with better ratings?\\
        \textbf{T5:} The Cityroomz has mixed reviews about their water pressure. Some guests found the water pressure to be very low while others found it to be disappointing. Would you like to know more about them?\\
        \textbf{T5$_\text{ABSA}$:} The Cityroomz has a lot of reviews that say the water pressure is very low and disappointing. Do you want to look at some other places?\\\noalign{\vskip 0.75ex}\hdashline\noalign{\vskip 0.75ex}
        \textbf{Reference:} \\
        No, guests consistently complain about the water pressure, unfortunately. Will that be okay or should I do another search?\\\bottomrule
        
    \end{tabular}
    \caption{\label{tab:sample} Sampled output of different RG models. }
\end{table}

\vspace{6pt}
\noindent \textbf{Qualitative Analysis}\quad
Table \ref{tab:sample} shows an example of responses generated by various systems.
In this example, all the reviews express negative opinions about water pressure. However, responses generated by GPT-2 and BART include positive opinions. T5 correctly mentions the negative opinions but the generated response is not natural and coherent. By incorporating the ABSA model, both BART and T5 correctly generate responses with all negative opinions. 

\begin{table}[t]
    \centering
    \small
    \tabcolsep 3.5pt
\renewcommand{\arraystretch}{1.2}
    \begin{tabular}{lccccccc}
    \toprule
        ~ & \multicolumn{2}{c}{\textbf{KS}} && \multicolumn{3}{c}{\textbf{RG}}   \\ %
        \cline{2-3} \cline{5-7}
        ~ & \textbf{Macro-F} & \textbf{mAP} && \textbf{BLEU} & \textbf{R-L} & \textbf{BS}  \\ \midrule
        RG & -  & - && 10.80 & 28.52 & 41.12  \\ %
        +KS & 84.60 & 91.84 && 10.20 & 27.78 & 40.64 \\ %
        +ET+KS & 83.47 & 90.45 && 10.29 & 27.80 & 40.56 \\ %
        +KTD+ET+KS & 83.46 & 90.45 && 10.27 & 27.79 & 40.55 \\ \bottomrule
    \end{tabular}

    \caption{\label{tab:pipeline_results} Results of the end-to-end evaluation. We start from RG with gold knowledge as input. We then gradually add components (KS, ET, and KTD) to the pipeline to replace the gold input with the predicted one. }
\end{table}

\begin{table}[t]
    \centering
    \small
    \tabcolsep 2.7pt
    \begin{tabular}{lcccccccc}
    \toprule
        ~ & \textbf{KTD} && \multicolumn{2}{c}{\textbf{KS}} && \multicolumn{3}{c}{\textbf{RG}}   \\ %
        \cline{2-2} \cline{4-5} \cline{7-9}
\noalign{\vskip 3pt} %

        ~ & \textbf{Acc} && \textbf{Macro-F} & \textbf{mAP} && \textbf{BLEU} & \textbf{R-L} & \textbf{BS}  \\ \midrule
        Fact-TOD & 87.62 && 59.55  & 76.69 && 6.15 & 23.25 &  33.16 \\ %
        SK-TOD & 99.71 && 84.60 & 91.84 && 10.80 & 28.57 & 41.12 \\ \bottomrule
    \end{tabular}

    \caption{\label{tab:dstc9_results} Comparison between models trained on Fact-TOD and SK-TOD training data. }
\end{table}

\section{Experiments on End-2-End Evaluation}
In the previous section, we use gold information as input for each module to avoid error accumulation. 
In this section, we evaluate the entire pipeline in an end-to-end manner, where the input of each subtask is predicted by the previous component. We gradually add KS, ET, and KTD to the pipeline, and list the performance of KS and RG in Table \ref{tab:pipeline_results}. 

The results show that errors introduced during KS can decrease the quality of response generation. However, ET and KTD do not have a significant impact on the performance of downstream tasks. It is because ET and KTD results include fewer noisy predictions compared to the KS results.

\section{Comparison with Fact-TOD}

One difference between SK-TOD and Fact-TOD is that responses in SK-TOD are grounded on subjective knowledge instead of factual knowledge. 
In this section, we investigate whether a Fact-TOD model can ground on subjective knowledge to address subjective requests. 
To this end, we re-train our KTD (DeBERTa), KS (DeBERTa cross-encoder), and RG (BART) models using the FAQ-grounded TOD data provided by \citet{kim-etal-2020-beyond}. We then apply it to the test set of SK-TOD without further training. 
We compare the results of each sub-task using the Fact-TOD models with the results of models trained on SK-TOD. 

As shown in Table \ref{tab:dstc9_results}, for all tasks, there is a significant performance gap between models trained on Fact-TOD and on SK-TOD. We also provide sampled outputs by the Fact-TOD model and the SK-TOD model in Table \ref{tab:sample_fact_all}.
By checking the model output, we observe that the Fact-TOD model tends to only ground on and copy information from a single knowledge snippet.
This behavior indicates that it is difficult to apply the Fact-TOD model to the SK-TOD task directly, as the Fact-TOD model lacks the ability to effectively aggregate information from multiple knowledge snippets, especially when there are diverse and contradictory opinions.
The results also highlight that compared to Fact-TOD, SK-TOD faces new challenges in terms of subjective content understanding and dialogue modeling when integrating subjective knowledge into the responses.

\begin{table}[t]
    \centering
    \scriptsize
    \centering
    \begin{tabular}{p{0.95\linewidth}}
    \toprule

\textbf{Dialogue History:}\\
        U: I am looking for a guesthouse to stay at that has free wifi.\\
        S: Ok.  Were you hoping for a particular price range or part of town?\\
        U: I don't care about the part of town or the price range, but it needs to be a three star hotel.\\
        S: The Hamilton Lodge would be perfect! It is a moderately priced 3 star guesthouse on the north side of town. Does that sound okay to you?\\
        U: Does it have comfortable beds?\\
        \noalign{\vskip 0.75ex}\hdashline\noalign{\vskip 0.75ex}
        \textbf{Reviews:}\\
        Review 1: The bed really needed a new mattress, too.\\
        Review 2: That being said, the room was spacious and the bed was very comfortable.\\
        Review 3: We both slept extremely well every night we were there.\\\noalign{\vskip 0.75ex}\hdashline\noalign{\vskip 0.75ex}
        \textbf{Model Outputs:}\\
        \textbf{Fact-TOD:} Yes, the Hamilton Lodge has a new mattress. Would you like to make a reservation?\\
        \textbf{SK-TOD:} The Hamilton Lodge has really comfortable beds according to most guests, but one guest did say that the bed needed a new mattress.\\
\noalign{\vskip 0.75ex}\hdashline\noalign{\vskip 0.75ex}
\textbf{Reference:} \\
There are some mixed reviews on the beds. Some say they're very comfortable while others were not impressed. Would you like me to find another place with better reviews?
        \\
        \bottomrule
        
    \end{tabular}
    \caption{\label{tab:sample_fact_all} Sampled outputs from the Fact-TOD model and the SK-TOD model, respectively. }
\end{table}

\section{Conclusion}
In this paper, we have introduced SK-TOD: a novel task focused on subjective-knowledge-based task-oriented dialogue response generation. 
We create and release a large-scale, manually-annotated dataset for this task. Incorporating subjective knowledge requires models to accurately identify all relevant knowledge snippets and faithfully aggregate the information into concise and contextually appropriate  responses, which brings unique challenges to this task. 
Experiments with strong baselines show that there is a significant performance gap between human-generated and machine-generated responses, particularly in faithfully capturing the diversity and proportion of opinions present in the subjective knowledge.
We hope this task together with the provided dataset can promote future research on knowledge-grounded TOD systems and subjective content understanding.

\section*{Limitations}

The dataset we collected contains two domains, restaurants and hotels. However, to evaluate the model's ability to generalize across different domains, it would be beneficial to include more domains in the dataset. Additionally, to address privacy and copyright concerns, we used crowd-sourcing to collect review data, resulting in fewer and shorter reviews than those found in real-world scenarios. This limitation can be mitigated by sampling informative and reliable reviews from real-world data. 
Regarding the model, we did not investigate more complex models, such as large language models and novel architectures. However, we provide a strong baseline method that will serve as a benchmark for more advanced methods by the research community.

\section*{Ethical Considerations}
To build our dataset, we collected the dialogue data by augmenting MultiWOZ 2.1, which is a publicly available English dialogue dataset under MIT license. Additionally, we collected the review data using crowd-sourcing, where we provided crowd workers with the reviewer’s persona, as well as the aspects and sentiments of reviews. This controlled review collection process helps to exclude offensive or harmful content from the reviews. It also helps to avoid privacy or copyright issues when making the dataset publicly available. Our dataset is available under the CDLA-Sharing 1.0 license. %

\bibliography{custom}

\appendix

\section{Data Collection}
\label{app:data}

In this section, we describe more details of the data collection process. 
The data collection is conducted by a group of crowd workers through Amazon Mechanical Turk. 
To control the data quality, 
we choose English speakers from the US, CA, and GB.
Workers are eligible for the annotation only if they pass our pre-qualification tests. During data collection, we also manually validate the annotation quality in several rounds to filter out the workers with low-quality annotations. 

\begin{figure}[t]
    \centering
    \includegraphics[width=\linewidth]{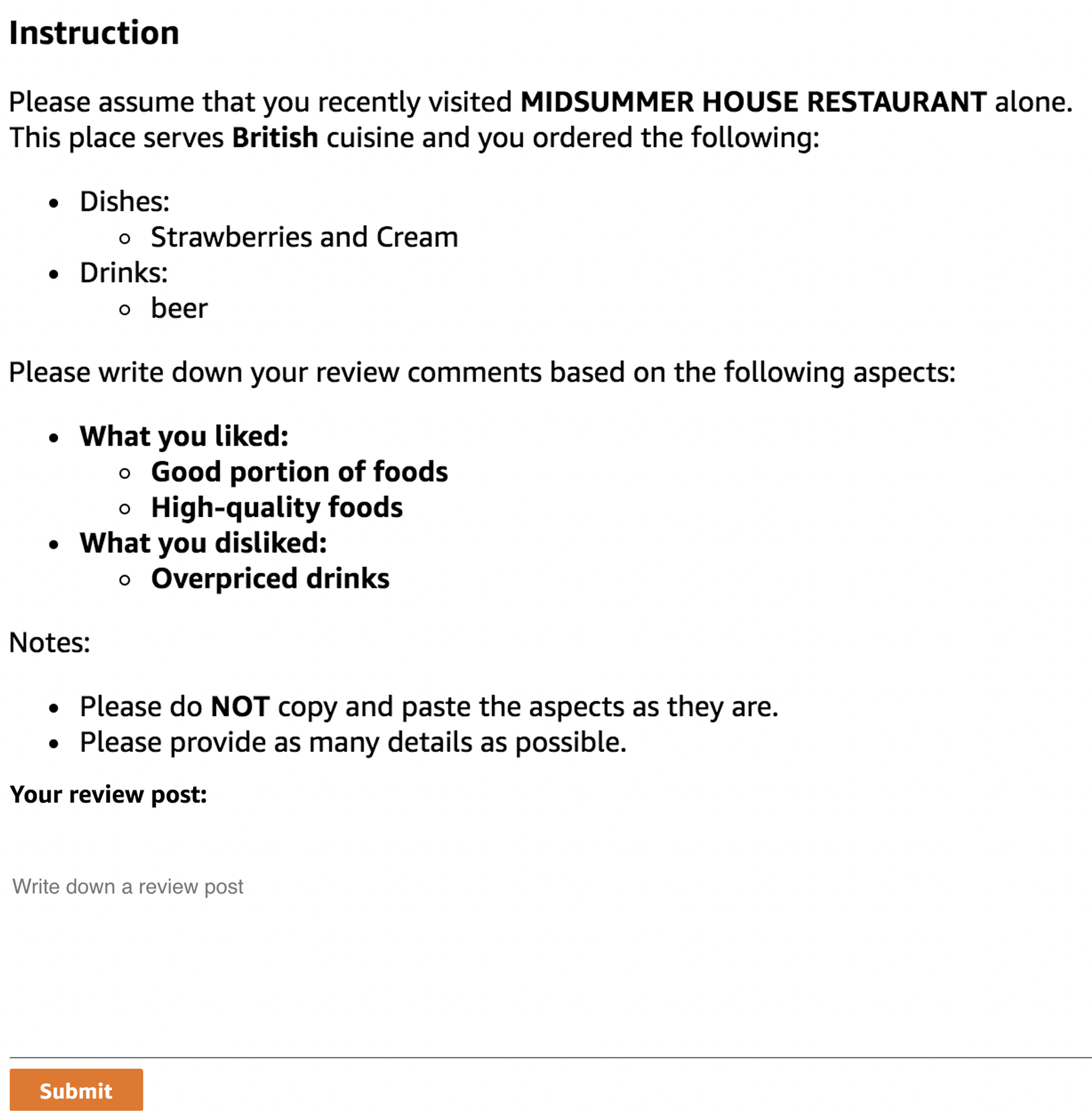}
    \caption{The interface of review collection.} %
    \label{interface:review}
\end{figure}

During review collection, we provide the reviewer's persona, as well as the aspects and sentiments of reviews to workers. The persona is randomly sampled from a pre-defined set of personas. For the aspects and sentiments, we first define 26 common aspects for hotel and restaurant reviews (e.g., WIFI-quality and room-bed for hotels, food-quality and indoor-decor for restaurants). We then randomly selected the target aspects to be addressed in a review. The number of aspects is randomly chosen. To mimic the sentiment distribution of the real reviews, the sentiment of each aspect is sampled based on the actual average ratings taken from Yelp. Figure \ref{interface:review} shows the interface of review collection. We pay workers \$1.00 per task.

\begin{figure}[t]
    \centering
    \includegraphics[width=\linewidth]{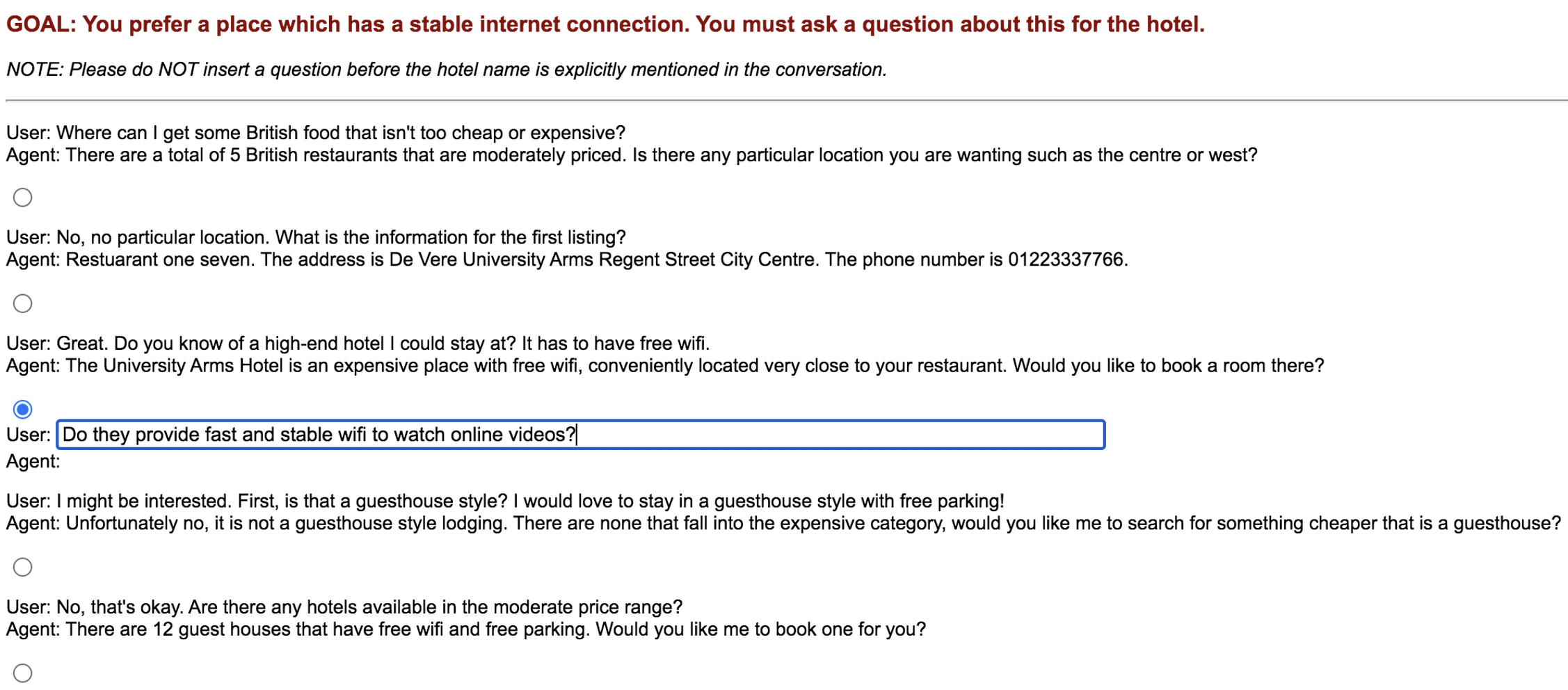}
    \caption{The interface of user request collection.} %
    \label{interface:request}
\end{figure}

During user request collection, we ask workers to select the best position to insert a user request by considering every possible position of the given dialogue. Figure \ref{interface:request} shows the interface of user request collection. We pay workers \$0.15 per task.

\begin{figure}[h]
    \centering
    \includegraphics[width=\linewidth]{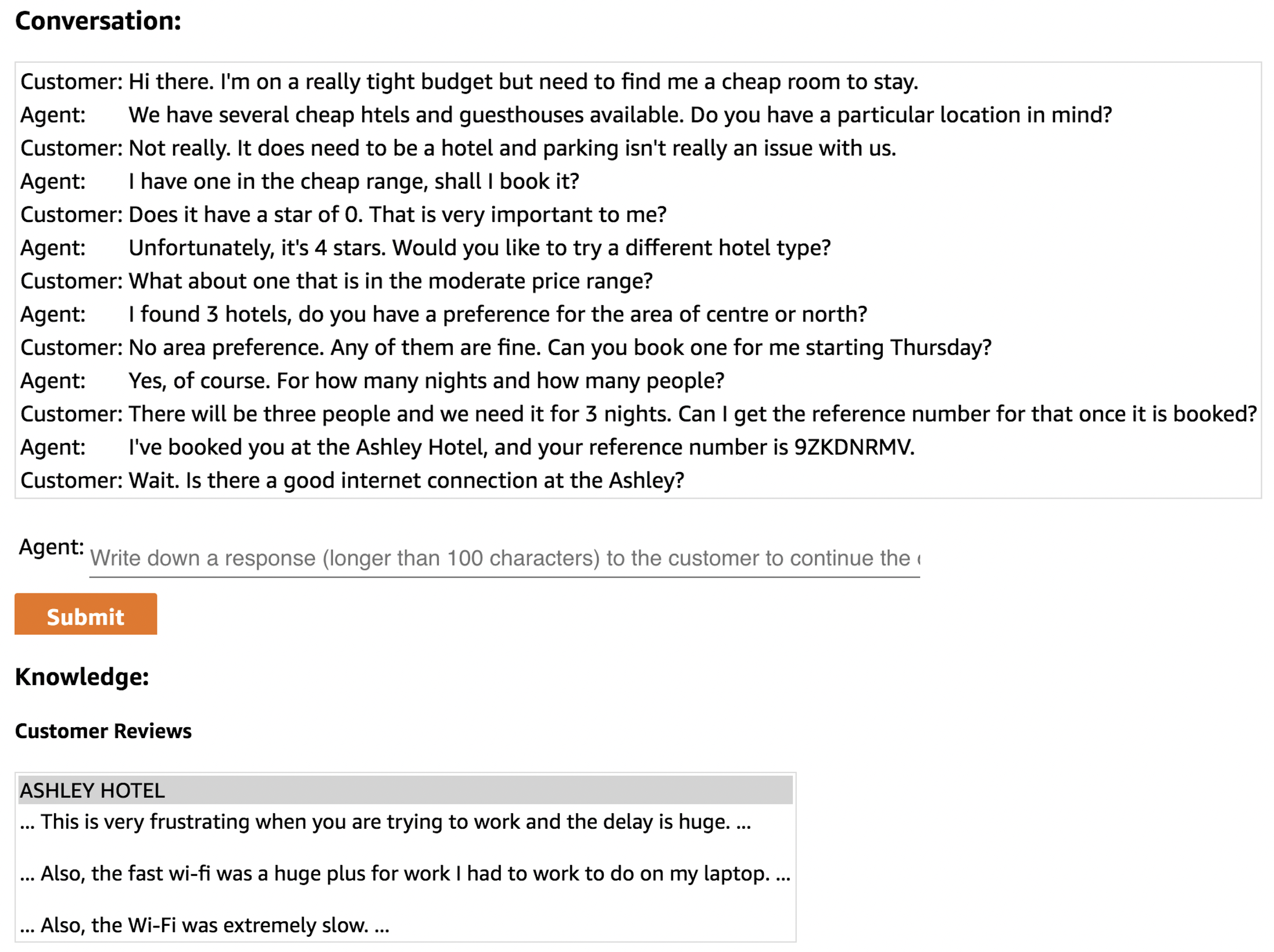}
    \caption{The interface of response generation.} %
    \label{interface:response}
\end{figure}

During response generation, we explicitly ask workers to consider the information in all
snippets to create a natural and faithful response. Figure \ref{interface:response} shows the interface of response generation. We pay workers \$0.25 per task. Below we list the complete instructions that we provide to workers. 
\begin{itemize}
\item Please read ALL the customer reviews carefully.
\item Please read the conversation carefully.
\item Write down a response to the customer to answer the question and continue the conversation.
\item You must read EVERY REVIEW COMMENT carefully. Each sentence was written by different people with potentially different opinions.
\item Your response MUST include your SUMMARY of ALL the review sentences.
\item If there's any conflict or different opinions in the reviews, your response MUST describe the minority opinion as well.
\item Your response MUST be based on the contents in given review comments only.
\item Please keep the way of speaking as similar as possible to the previous utterances spoken by the agent.
\end{itemize}

\section{Aspect Based Sentiment Analysis}
\label{app:absa}
To enhance the model's ability to understand the sentiment polarity of each individual knowledge snippet, we apply \textsc{PGen} \cite{zhang-etal-2021-aspect-sentiment}, a state-of-the-art aspect-based sentiment analysis model, to predict the sentiment $Z=[z_1, z_2, \cdots, z_i, \cdots]$ for every knowledge snippet $[K_1, K_2, \cdots, K_i, \cdots]$ in $\mathcal{K}^+$. 

\textsc{PGen} converts the problem of aspect-based sentiment analysis into a sequence generation problem, where the input is the review sentence, and the output is a natural language description of the aspect and the sentiment. For example, given the review sentence as ``\textit{The ambience was so fun.}'', where the aspect term is ``ambience'' and the corresponding sentiment polarity is ``positive'', \textsc{PGen} transform the aspect term and the sentiment polarity into a natural language description ``ambience is great.'' using templates. It is transformed by keeping the aspect term unchanged and mapping the positive/neutral/negative sentiment polarities into one of the three tokens: ``great'', ``ok'', and ``bad''. The model is trained using a BART-base model on semeval aspect-based sentiment analysis datasets \cite{pontiki2015semeval,pontiki2016semeval}.

\section{Training Details}
\label{app:training}
For KTD and KS, the implementation is based on Transformers \cite{wolf-etal-2020-transformers}. 
During training, we use AdamW \cite{loshchilov2018decoupled} with a learning rate of $3 \times 10^{-5}$ and a batch size of 16. We apply warmup \cite{goyal2017accurate} on the first 500 steps and early stopping based on the model performance on the validation set. We use a Tesla V100 GPU with 16 GB memory for training models. It takes 1 hour to train a KTD model and 5 hours to train a KS model. 

During inference, we set the classification threshold as 0 for KTD, as we observe that KTD results are insensitive to the threshold. However, for the KS model, the setting of the threshold can greatly impact the precision and recall scores. We therefore choose the best threshold based on the $F_1$ scores on the validation set. We use a grid search between -5 to 5. The optimal thresholds for BERT, RoBERTa, ALBERT, and DeBERTa are 2.25, 1, 1.75, and 2 in the bi-encoder setting. They are 3.1, 4.6, 3.25, and 3.4 in the cross-encoder setting.  

For ET model, we follow the setting of \citet{jin2021can} to identify entities. More specifically, we perform the fuzzy n-gram
matching between an entity and the utterance, where n is the same as the length of the entity mention. The n-gram
matching score is calculated based on the ratio of the longest common sequence between two n-grams. We set the matching threshold as 0.95. 

For RG model, during training, we use AdamW with a learning rate of $3 \times 10^{-5}$ and a batch size of 16. 
We apply the warmup on the first 500 steps and the early stopping based on the model performance (perplexity) on the development set. The model is trained on a Tesla V100 GPU with 16 GB memory for 2 hours.

\end{document}